\documentclass[sn-mathphys]{sn-jnl}
\usepackage{color}
\usepackage{marvosym}

\jyear{2021}%

\theoremstyle{thmstyleone}%
\theoremstyle{thmstyletwo}%

\theoremstyle{thmstylethree}%

\begin{document}

\title[Article Title]{Cross Attention-guided Dense Network for Images Fusion}


\author{\fnm{Zhengwen Shen}}\email{shenzw21@cumt.edu.cn}

\author{\fnm{Jun Wang\textsuperscript{\Letter}}}\email{\textsuperscript{\Letter}wj999lx@cumt.edu.cn}

\author{\fnm{Zaiyu Pan}}\email{pzycumt@163.com}
\author{\fnm{Yulian Li}}\email{lylcumtedu@163.com}
\author{\fnm{Jiangyu Wang}}\email{wongjoyu@163.com}
\affil{\orgdiv{School of Information and Control Engineering}, \orgname{China University of Mining and Technology}, \city{Xuzhou}, \postcode{221000}, \country{China}}




\abstract{In recent years, various applications in computer vision have achieved substantial progress based on deep learning, which has been widely used for image fusion and shown to achieve adequate performance. However, suffering from limited ability in modeling the spatial correspondence of different source images, it still remains a great challenge for existing unsupervised image fusion models to extract appropriate feature and achieves adaptive and balanced fusion. In this paper, we propose a novel cross-attention-guided image fusion network, which is a unified and unsupervised framework for multi-modal image fusion, multi-exposure image fusion, and multi-focus image fusion. Different from the existing self-attention module, our cross-attention module focus on modeling the cross-correlation between different source images. Using the proposed cross attention module as a core block, a densely connected cross attention-guided network is built to dynamically learn the spatial correspondence to derive better alignment of important details from different input images. Meanwhile, an auxiliary branch is also designed to capture more context information, and a merging network is attached to finally reconstruct the fusion image. Extensive experiments have been carried out on publicly available datasets, and the results demonstrate that the proposed model outperforms the state-of-the-art quantitatively and qualitatively. The source code is available at https://github.com/shenzw21/CADNIF.}

\keywords{deep learning, image fusion, cross attention, spatial correspondence}



\maketitle

\section{Introduction}\label{sec1}

Image fusion aims at unifying the feature information from different source images into an efficient representation and advancing the visual perception performance of humans and machines. Particularly, image fusion has a wide and various applications in computer vision, such as multi-modal image fusion: infrared and visible image fusion\cite{li2018densefuse}, RGB and depth image fusion for semantic segmentation\cite{wang2020deep}, medical image fusion\cite{ma2020ddcgan}, etc; multi-focus image fusion\cite{zhang2021mff}; multi-exposure image fusion\cite{ram2017deepfuse}, etc.
\\ \indent In recent years, traditional methods for image fusion take the lead and group three main directions: transform domain approach\cite{cao2014multi}, sparse representation\cite{bin2016efficient,zhang2018sparse}, various component analysis\cite{kuncheva2013pca}. While many traditional methods are proposed and perform well in image fusion task, there also exists shortcoming and limitations: the feature extraction relies on handcrafted, and lack flexibility and generalization ability. With the development and wide application of deep learning, the limitations of traditional methods got a certain breakthrough and many image fusion methods based on deep learning are proposed\cite{ram2017deepfuse,li2018densefuse,ma2019fusiongan}. Although existing methods proposed based on deep learning improve the performance and generalization ability, image fusion still suffer from feature extraction and modeling spatial correspondence learning from source images issues. For example, the design fusion strategy needs more artificial for different fusion tasks, the balance of feature extraction, and effective joint representation learning from source images.
\\ \indent In order to solve the mentioned problems, in this paper, we proposed a novel cross-attention-guided dense network for image fusion which is constructed by cross attention guided densenet, auxiliary network, and merging network. Modeling the cross-correlation, we proposed the cross-attention-guided dense network focus on the spatial correspondence from source images by a densely connected strategy. Avoiding a lack of global information, we adopted an auxiliary network to build the long-distance correspondence for source images. To better reconstruct the fusion image, we adopted a residual connect merging network to aggregate the fusion feature from source images.
\\ \indent \noindent The main contributions of this paper are summarized as follows:
\begin{itemize}
\item A. We propose a novel multi-task images fusion method, which is named Cross Attention-guided Dense Network for Images Fusion(CADNIF). It utilizes the advantage of the deep neural network model with attention mechanisms and improves the ability of feature extraction and modeling spatial correspondence from source images effectively and in robustness.
\item B. We introduce different cross-attention-guided fusion modules to effectively model the spatial correspondence and global correspondence from source images, the cross spatial attention, and the cross self-attention are applied according to the characteristics in the source images.
\item C. Extensive experiments on different datasets validate the superiority of the proposed CADNIF, which is successfully applied to multi-image fusion tasks with superior performance against the state-of-the-art methods.
\end{itemize}

\section{Related Work}
In this section, we group the image fusion methods into two categories, traditional methods and deep learning-based approaches for image fusion. We also briefly introduce the dense block and attention mechanisms, which are highly related to our proposed method.

\subsection{Image fusion.}
In the field of image fusion, the key to the evaluation of fusion algorithms is how to effectively extract features and fuse features. Numbers of traditional image fusion methods have been proposed to solve the feature extract problem, and group the methods into three main categories: transform domain approach, such as discrete cosine(DCT), discrete wavelet(DWT), etc; sparse representation domain approach; various component analysis, such as principal component analysis(PCA), independent component analysis(ICA), etc. However, the feature extraction methods lack flexibility and generalizability because of the increasing complexity. Moreover, traditional methods need to pay much attention to designing the appropriateness of fusion methods to ensure the features come from different specific source images.
\\ \indent With the wide application of deep learning in high-level vision tasks, the limitations of traditional methods have a breakthrough to a certain extent. More and more CNNs based approaches were introduced for feature extraction as a backbone for image fusion tasks via various fusion strategies. The deep learning-based approaches for image fusion have been application successful in several areas, such as multi-modal image fusion, multi-exposure image fusion, multi-focus image fusion, and medical image fusion. \cite{ram2017deepfuse} proposed a deep learning-based unsupervised Deepfuse framework for exposure and extreme exposure image fusion. An infrared and visible image fusion method named DenseFuse based on Deepfuse and incorporated dense block in the encoding network for feature extraction is proposed\cite{li2018densefuse}. FusionGAN is based on the generative adversarial network for infrared and visible image fusion\cite{ma2019fusiongan}. \cite{ma2020ddcgan} extended previous work to adopt a dual-discriminator strategy for fusing multi-modality images of multi-resolution image fusion. Recently, some works are focused on unified unsupervised end-to-end image fusion networks, such as multi-modal multi-exposure, multi-focus\cite{xu2020u2fusion,zhang2020rethinking}. However, the above methods focus on the significance of the difference between the different source images and do not consider the adaptability balance of the final fusion information from source images. In this paper, we propose a more generic unified unsupervised approach to multi-image fusion task, which joint learning spatial correspondence information from source images, and for each cross attention module outputs, makes it possible for building a balance that each source image becomes one.

\subsection{Image fusion with dense block.}
While deep convolution neural network has achieved remarkable results in the field of computer vision, with the wide application and the depth increased, gradually exposed a lot of problems, such as performance degradation, vanishing gradient, exploding gradient, and previous features not reused fully, etc. To address the above problems, a deep residual learning network is proposed\cite{he2016deep}. Based on the residual work, \cite{huang2017densely} proposed a new network framework densenet, in which each layer can be concatenated by the previous layer and as an input for the next layer and strengthen feature propagation, encourage feature reuse, and alleviates the problem of the gradient. Based on the above advantages, the dense block has been widely incorporated into multi-computer vision tasks, such as semantic segmentation, image classification, object detection, etc. In the image fusion task, a dense block is widely used in feature extraction. \cite{li2018densefuse} incorporate the dense block into the encoding network, preserving the useful information from the middle layers. \cite{ma2020ddcgan} proposed a novel GAN framework for image fusion and the architecture of the generator is based on densenet. Instead of directly using densenet as a feature extraction network, we propose a cross-attention-guided dense network, which effective learning cross-correlation information between different source images.
\subsection{Attention mechanisms in deep learning methods.}
In recent years, with the emergence of the attention mechanism, which is a technique that enables models to focus on important information and fully learn to absorb it, it has been widely used in many computer vision tasks, natural language processing, image, and video understanding. A novel channel attention network for multi-spectral imagery\cite{9025565}.  To address the problem of Crowd Counting, multi-level attention is proposed\cite{tian2021multi}. \cite{song2021cross} proposed cross-modal attention for MRI and Ultrasound Volume Registration. Our work explores cross attention for basic mutual feature extraction of source images.
\begin{figure*}[t]
\centering
\includegraphics[width=0.8\textwidth]{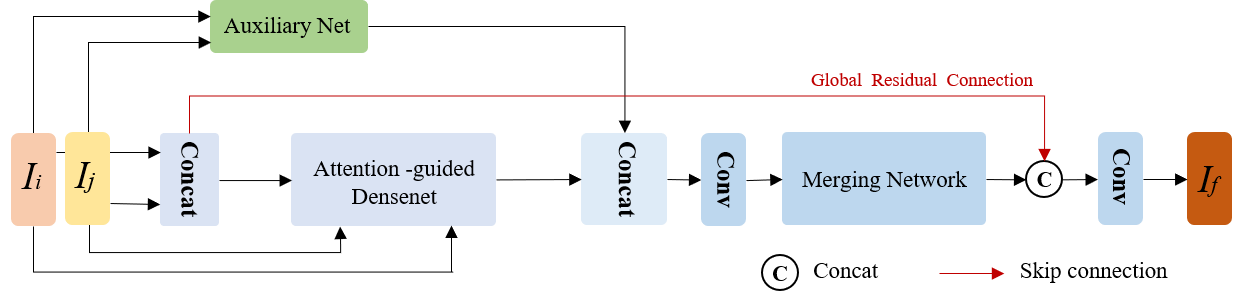} 
\caption{The overall architecture of the proposed CADNIF mainly consists of three components: Attention-guided Network, Auxiliary network, and Merging Network. Firstly, the attention-guided network is adopted to extract cross-attention features from source images. Simultaneously, the auxiliary network extracts the long-range feature representations via cross self-attention. Finally, the merging network merges the global feature representations and the spatial correspondence via a dilated residual dense block and global residual connection.}
\label{fig1}
\end{figure*}

\section{Methods}
We propose a novel cross-attention-guided image fusion framework, which focuses on improving the cross-significance information representation, modeling the spatial correspondence, capturing more context information, and merging the feature from source images to reconstruct the fusion image. As shown in Figure~\ref{fig1}, the structure consists of three networks: cross attention-guided dense network, auxiliary network, and merging networks. The cross attention-guided network captures the spatial correspondence feature via using five attention blocks based on densenet, the responsibility of the auxiliary for capturing more context information, while the merging network relies on a series of dilated residual dense blocks to utilize the image features effectively and adopt a global residual connect to obtain more detail information. Finally, the proposed attention-guided fusion network further to fuse the source images and the following experiments show that the fusion result prefers well.
\subsection{Cross Attention Guided Densenet}
\begin{figure*}[ht!]
\centering
\includegraphics[width=0.8\textwidth]{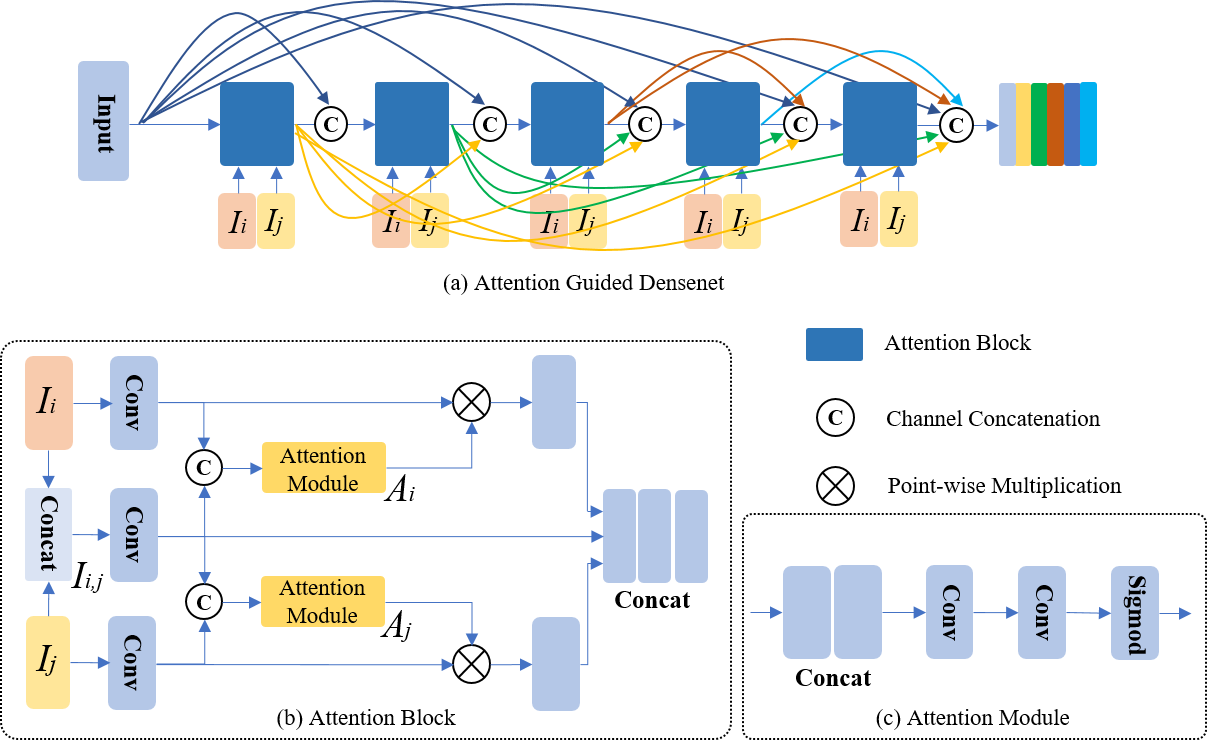} 
\caption{(a) The architecture of the proposed attention-guided dense network, network consists of five attention blocks for modeling the spatial correspondence. (b) shows the cross attention blocks. (c) shows the attention module.}
\label{fig2}
\end{figure*}

Unlike the previous methods, which use the pure densenet as a feature extract network directly, our proposed cross-attention guided densenet obtains the cross-attention information from each source image. As shown in Figure~\ref{fig2}, the attention-guided densenet consists of five major cross-attention blocks. Given input images $I_{i}$ and $I_{j}$, each input image is concatenated by three one-channel gray images of the same source. We first concatenate the input images and obtain the 6-channel concatenation image. As shown in Figure~\ref{fig2}a, the input images and the concatenated image feed into the attention-guided network, and obtain the final cross-attention features via dense connection. As shown in Figure~\ref{fig2}b, the cross attention guided block consists of three convolution layers to obtain feature maps, concatenate operation, and attention module. Details of the cross attention block are provided below. The final cross-attention maps can be obtained via:
\begin{equation}\label{eq1}
\begin{split}
 A_{i}= Attent\left ( I_{i}, Concat\left ( I_{i}, I_{j} \right ) \right )
\end{split}
\end{equation}
\begin{equation}\label{eq2}
\begin{split}
 A_{j}= Attent\left ( I_{j}, Concat\left ( I_{i}, I_{j} \right ) \right )
\end{split}
\end{equation}
\begin{equation}\label{eq3}
\begin{split}
 Z_{g}= Concat\left ( Concat\left ( I_{i}, I_{j} \right ), A_{i}\circ I_{i}, A_{j}\circ I_{j} \right )
\end{split}
\end{equation}
where $\circ$ denotes the point-wise multiplication, $I_{i}$ and $I_{j}$ denote the original input images, $A_{i}$ and $A_{j}$ denote the attention map, $Z_{g}$ denote the final cross attention feature maps obtained from source images.

\subsection{Attention Module}
Attention modules play an important role in the cross-dense attention block, as shown in Figure~\ref{fig2}c, to obtain the one-to-one correspondence maps for each source image, the attention module consists of two convolution layers and a Sigmoid layer. Each convolution layer applies $3\times 3$ layers. Respectively, a ReLU activation and a Sigmoid activation in the module after convolution operation. As a result, we finally obtain the attention map $A$ with values in the range $\left [ 0, 1 \right ]$.
\subsection{Auxiliary network}
\begin{figure}[ht]
\centering
\includegraphics[width=0.5\columnwidth]{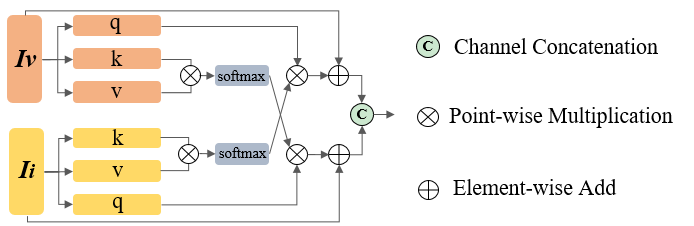} 
\caption{The architecture of the auxiliary network.}
\label{fig3}
\end{figure}

As shown in Figure~\ref{fig3}, in the auxiliary network, we first use $1\times 1$ convolution kernel to obtain 32-channels feature maps via concatenating operation of the input images $I_{i}$ and $I_{j}$. To aggregate the features and obtain the long-range information, we propose a cross self-attention module by considering the relation between source images. The cross self-attention operation is defined as follows:
\begin{equation}\label{eq4}
\begin{split}
 y_{i,j}= \frac{1}{C\left ( x \right )}\sum_{\forall i,j}^{}f\left ( f_{i},f_{j} \right )g\left ( f_{j} \right )
\end{split}
\end{equation}
where $f_{i}$ and $f_{j}$ denote the infrared and visible image, respectively. Function f produces the adaptive pixel level weight vectors between source images and function g produce a feature representation of the input single image. The normalization factor is defined ss ${C\left ( x \right )}= \sum_{\forall i,j}^{}f\left ( f_{i},f_{j} \right )$.

\subsection{Merging Network}
\begin{figure}[ht]
\centering
\includegraphics[width=0.4\columnwidth]{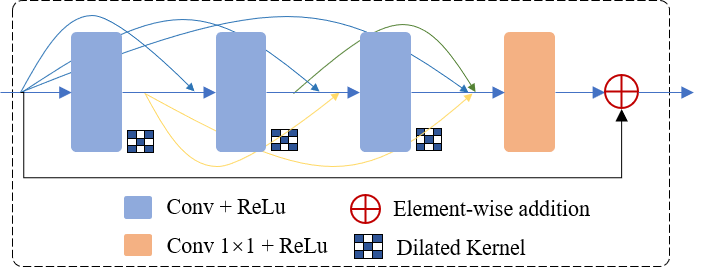} 
\caption{The detailed architecture of the merging network.}
\label{fig4}
\end{figure}

To better reconstruct the fused image, and integrate the cross-correlation information and long-range information to obtain more details of the feature maps, we adopt a dilated residual dense block proposed in~\cite{yan2019attention} as the reconstructed network. The merging network takes the concatenated feature maps after a convolution layer of $1\times 1$ kernel operation as input, which is concatenated from the attention-guided network and auxiliary network. As shown in figure~\ref{fig4}, the dilated residual dense block consists of three convolution layers followed by ReLu activation, concatenation-based skip-connection similar to densenet, and a convolution layer of $1\times 1$ kernel operation as output. Different from the normal convolution operation, each convolution layers adopt 2-dilated convolutions, and kernel size is $3\times 3$. Finally, we apply a global residual connection strategy to concatenate the output with concatenation images from source images.
\subsection{Training Loss}
As described above section, the proposed CADNIF method can be trained to obtain the context details structural and background details from the different sources images. In the proposed method, the loss function consists of Mean Square Error($MSE$) loss $L_{mse}$ and gradient loss $L_{grad}$.
The gradient loss forces the fused image to contain rich texture details, the $L_{grad}$ is defined by Eq.~\ref{eq4},
\begin{equation}\label{eq4}
\begin{split}
 L_{grad}=\frac{1}{HW}\left\|\triangledown I_{fused}-I_{original} \right\|_{2}^{2}
\end{split}
\end{equation}
where $I_{original}$ denotes the input source image, $I_{fused}$ denotes the fused image ,$\triangledown$ denotes the gradient operator, H and W are the height and width of the image respectively, $\left \| \cdot  \right \|_{2}^{2}$ denotes the $l_{2}$ normal.
The $MSE$ loss emphasizes the matching of each corresponding pixel between the input image and output image, the $MSE$ loss $L_{mse}$ is calculated as Eq.~\ref{eq5},
\begin{equation}\label{eq5}
\begin{split}
 L_{mse}= \frac{1}{HW}\left\|I_{fused}-I_{original} \right\|_{2}^{2}
\end{split}
\end{equation}
where $I_{original}$ denotes the input source image, $I_{fused}$ denotes the fused image.
Finally, the total losses $L_{total}$ of the proposed model can be expressed as follows:
\begin{equation}\label{eq6}
\begin{split}
L_{total}= \lambda _{mi}I_{mse_i}+\lambda _{mj}I_{mse_j} + \lambda _{gi}I_{grad_i}+\lambda _{gj}I_{grad_j}
\end{split}
\end{equation}
where $\lambda _{\left ( \cdot  \right )}$ denotes the contribution of each loss to the whole objective function. In this paper, for each experiment below, setting the parameter $\lambda _{mi}$, $\lambda _{mj}$, $\lambda _{si}$, $\lambda _{sj}$ equal to: 1, 1, 4, 4, in infrared and visible fusion experiment; 0, 1, 1, 0, in MRI and PET fusion experiment; 0.0005, 0.0005, 1, 1, in Multi-focus fusion experiment; 1, 1, 4, 4, in Multi-exposure fusion experiment.
\section{Experiments}
To demonstrate the proposed method, we conduct the experiments on five publicly available datasets: TNO$\footnote{http://figshare.com/articles/TNOImageFusionDataset/1008029}$ and RoadScene$\footnote{https://github.com/hanna-xu/RoadScene}$ datasets for visible and infrared image fusion tasks, TNO used for training, RoadSence are all used for testing; MFF and SICE datasets$\footnote{https://drive.google.com/drive/folders}$ for multi-exposure image fusion task; MRI(Magnetic Resonance Imaging) and PET(Positron emission
tomography)$\footnote{http://www.med.harvard.edu/AANLIB/home.html}$ datasets for medical image fusion task; Lytro$\footnote{https://github.com/sametaymaz/Multi-focus-Image-Fusion-Dataset}$ for multi-focus image fusion task. For the lack of training data problem, we adopt the expansion strategy based on\cite{zhang2020rethinking}, and finally obtain 20036, 33961, 4698, and 139001 pairs of training data, respectively. For the testing, the number of image pairs used for the testing dataset is 17, 50, 19, 20, and 18, respectively.
\\ \indent For all experiments, we set the batchsize, learning rate, and epoch equal to 16, $10^{-5}$, 10, respectively. The proposed method was implemented on NVIDIA GEFORCE RTX 2080 Ti GPU and based on TensorFlow.
\subsection{Infrared and visible image fusion}
\begin{figure*}[t]
\centering
\includegraphics[width=0.95\textwidth]{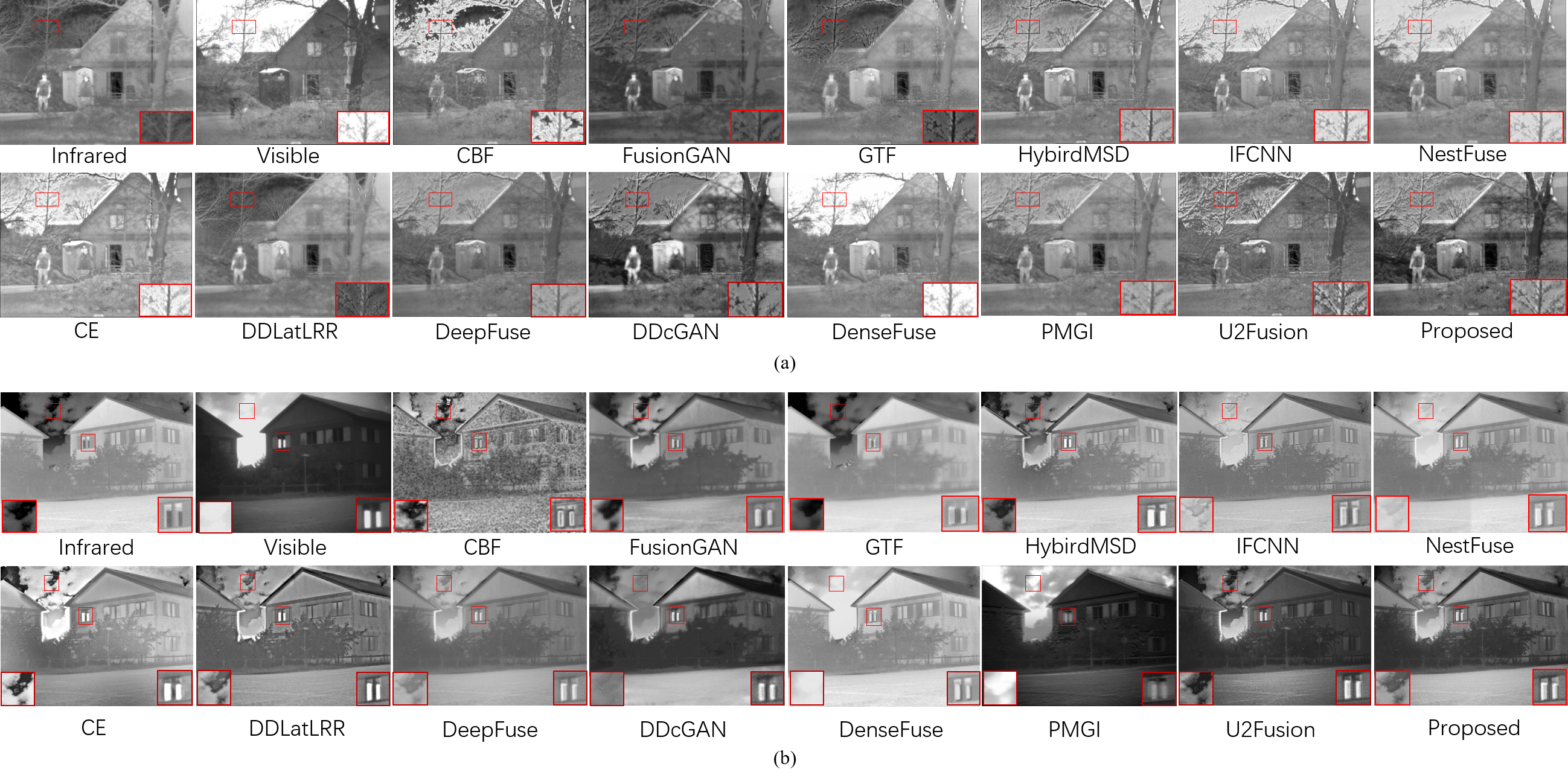} 
\caption{Comparison result of infrared and visible image fusion of TNO datasets. From left to right: infrared image, visible image, the results of CBF, FusionGAN, GTF, HybirdMSD, IFCNN, NestFuse, CE, DDLatLRR, DenseFuse, PMGI, U2Fusion, and proposed.}
\label{fig5}
\end{figure*}

\begin{figure*}[t]
\centering
\includegraphics[width=0.95\textwidth]{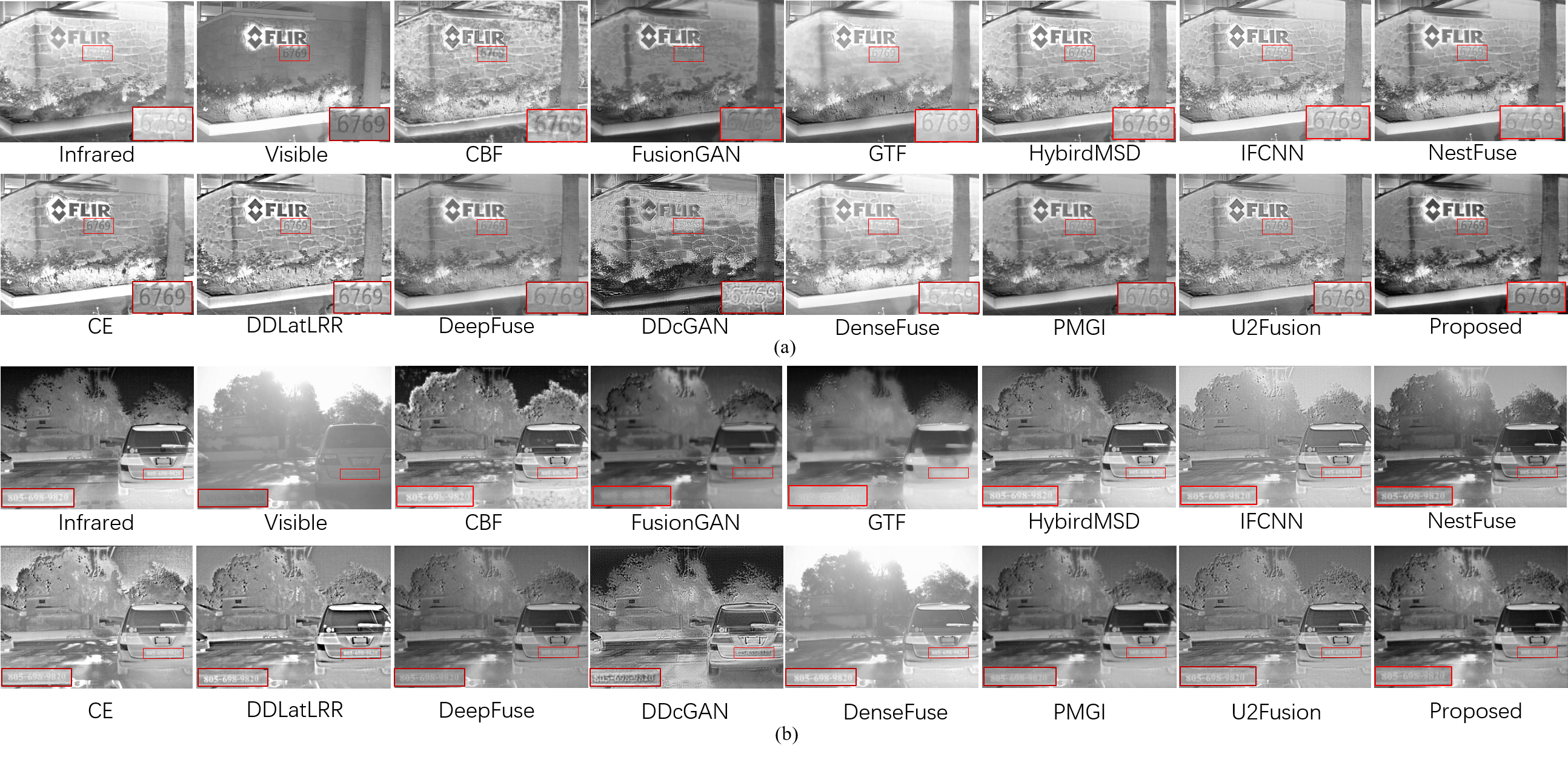} 
\caption{Comparison result of infrared and visible image fusion of RoadSence datasets. From left to right: infrared image, visible image, the results of CBF, FusionGAN, GTF, HybirdMSD, IFCNN, NestFuse, CE, DDLatLRR, DenseFuse, PMGI, U2Fusion, and proposed.}
\label{fig6}
\end{figure*}

In the infrared and visible image fusion experiment, most TNO datasets are used to train the proposed image fusion method, image pairs used for testing is 17, and 50 pairs of RoadSence datasets are used to test our proposed method to verify its generalization and model robustness. We compare our proposed method with CBF\cite{shreyamsha2015image}, FusionGAN\cite{ma2019fusiongan}, GTF\cite{ma2016infrared}, HybirdMSD\cite{naidu2014hybrid}, IFCNN\cite{zhang2020ifcnn}, NestFuse\cite{li2021rfn}, CE\cite{zhou2016fusion}, DDLatLRR\cite{li2020mdlatlrr}, DeepFuse\cite{ram2017deepfuse}, DDcGAN\cite{ma2020ddcgan}, DenseFuse\cite{li2018densefuse}, PMGI\cite{zhang2020rethinking}, U2Fusion\cite{xu2020u2fusion}.
\\ \indent To effective quantitatively evaluate the fusion quality, we adopt five related indicators to compare the existing methods, namely entropy (EN), the sum of the correlations of differences (SCD)\cite{aslantas2015new}, standard deviation (SD), represents the ratio of noise added to the final image(Qabf), and mutual information (MI). As shown in Table~\ref{table1} and Table~\ref{table2}, the quantitative result of TNO and RoadSence dataset, the red font in italic represent the best value and the bold black font in italic represents the second best value. We can infer that the quantitative value of the proposed method outperforms EN, SCD, SD, and MI metrics in the experiment of TNO dataset. Similarly, the quantitative value of the proposed method outperforms EN, SD, and MI metrics in the experiment of RoadSence dataset. It demonstrates that the proposed method maintains the abundant correspondence between spatial information and detailed information from source images.
\\ \indent The qualitative fusion results on two typical image pairs of each test dataset are illustrated in Figure~\ref{fig5} and Figure~\ref{fig6}, red box highlights are the local fine features. From the results of TNO dataset and RoadSence dataset, we can observe that the proposed method capture abundant correspondence semantic information compare with the state-of-the-art methods. Similarly, the highlights information obtained by zooming in on local fine features, we can infer that the local fine information is retained well from source images, such as sky, window, license plate, the top of the tree, etc. In particular, for each source image illumination information and texture details, the fusion result shows that the proposed method performs well by adaptive balance strategy via a cross attention-guided network.
\begin{table}[!h]
\renewcommand\tabcolsep{4.5pt}
\centering
\small\begin{tabular}{cccccc}
\hline
Methods & EN & SCD & SD & Qabf & MI \\
\hline
CBF & 6.9231 & 1.3880 & 81.5135 & 0.4328 & 13.8463 \\
FusionGAN & 6.5000 & 1.4992 & 63.1013 & 0.2163 & 13.0001 \\
GTF & 6.7804 & 0.9935 & 75.3037 & 0.4196 & 13.5610 \\
HybridMSD & 6.8094 & 1.6484 & 72.4155 & 0.5072 & 13.6188 \\
IFCNN & 6.7446 & 1.7340 & 74.1133 & 0.5074 & 13.4893 \\
NestFusion & 6.9810 & 1.7665 & 86.7143 & \textbf{0.5214} & 13.9620 \\
CE & 7.1939 & 1.5846 & 90.6729 & 0.4798 & 14.3879 \\
DDLatLRR & 6.8565 & 1.7255 & 76.3305 & 0.4881 & 13.7131 \\
DeepFuse & 6.8018 & \textbf{1.8508} & 74.1958 & 0.4392 & 13.6038 \\
DDcGAN & \textbf{7.2532} & 1.6927 & \textbf{102.5569} & 0.3519 & \textbf{14.4064} \\
DenseFuse & 6.8425 & 1.6183 & 89.3237 & \textcolor{red}{0.5350} & 13.6851 \\
PMGI & 7.1191 & 0.1062 & 99.6564 & 0.1535 & 14.1382 \\
U2Fusion & 6.8959 & 1.5877 & 76.4380 & 0.3941 & 13.7919 \\
Proposed & \textcolor{red}{7.4112} & \textcolor{red}{1.9044} & \textcolor{red}{103.5565} & 0.4070 & \textcolor{red}{14.8226} \\
\hline
\end{tabular}
\caption{Quantitative evaluation for each method via 17 pairs infrared and visible image fusion of TNO dataset.}\small
\label{table1}
\end{table}
\begin{table}[!h]
\renewcommand\tabcolsep{4.5pt}
\centering
\small\begin{tabular}{cccccc}
\hline
Methods & EN & SCD & SD & Qabf & MI \\
\hline
CBF & 7.5322 & 1.1477 & 82.3360 & 0.4793 & \textbf{15.0645} \\
FusionGAN & 7.1753 & 1.3752 & 67.0644 & 0.2723 & 14.3507 \\
GTF & \textcolor{red}{7.6346} & 1.1040 &84.7265 & 0.3787 & 14.2693 \\
HybridMSD & 7.4547 & 1.5757 & 78.7344 & \textcolor{red}{0.5425} & 14.9096 \\
IFCNN & 6.9730 & 1.5888 & 56.8367 & 0.5119 & 13.9460 \\
NestFusion & 7.3597 & 1.7564 & 77.7695 & 0.4856 & 14.7195 \\
CE & 7.3635 & 1.5893 & 75.8509 & 0.4587 & 14.7271 \\
DDLatLRR & 7.1485 & 1.6482 & 67.6889 & 0.5017 & 14.2971 \\
DeepFuse & 7.2034 & \textbf{1.8509} & 70.1455 & 0.4791 & 14.4069 \\
DDcGAN & \textbf{7.6088} & 0.8730 & \textbf{88.8762} & 0.2941 & 15.0177 \\
DenseFuse & 6.8549 & 1.2950 & 65.5098 & \textbf{0.5074} & 13.7099 \\
PMGI & 7.2492 & 1.7410 & 78.2691 & 0.4230 & 14.6985 \\
U2Fusion & 7.1968 & 1.7692 & 68.0393 & 0.4791 & 14.3937 \\
Proposed & 7.5437 & \textcolor{red}{1.9101} &  \textcolor{red}{89.0933} & 0.4549 & \textcolor{red}{15.0875} \\
\hline
\end{tabular}
\caption{Quantitative evaluation for each method via 50 pairs infrared and visible image fusion of RoadSence dataset.}\small
\label{table2}
\end{table}
\subsection{Unsupervised Multi-exposure Image Fusion}
In the multi-exposure image fusion experiment, we employ five existing methods to compare with our method: FMMR\cite{li2012fast}, DSIFT\cite{hayat2019ghost}, DeepFuse\cite{ram2017deepfuse}, MGFF\cite{bavirisetti2019multi} and IFCNN\cite{zhang2020ifcnn}, respectively.
\\ \indent From Table~\ref{table3}, compare the five existing methods by entropy (EN), Feature mutual information (FMI), standard deviation (SD), correlation coefficient (CC), and mutual information (MI). We can observe that the proposed fusion method outperforms EN, SD, MI, and CC metrics. Visual comparison results of each image fusion method can be seen in Figure~\ref{fig9}. From the perspective of global image information, the proposed method adaptive balances the illuminate information well, similarly the local context detail information is retained and presented from source images.
\begin{table}[!h]
\renewcommand\tabcolsep{4.5pt}
\centering
\small\begin{tabular}{cccccc}
\hline
Methods & EN & $FMI_{pixel}$ & SD & CC & MI \\
\hline
FMMR & 6.3172 & 0.8736 & 92.5337 & 0.5448 & 12.6344 \\
DISIFT & 6.5245 & \textcolor{red}{0.8934} & 101.3824 & 0.5445 & 13.0491 \\
DeepFuse & 6.8183 & \textbf{0.8926} & \textbf{116.3688} & 0.7945 & \textbf{13.8566} \\
MGFF & 6.8792 & 0.8874 & 116.2462 & \textbf{0.7976} & 13.7586 \\
IFCNN & 6.7259 & 0.8838 & 113.8987 & 0.7572 & 13.4519 \\
Proposed & \textcolor{red}{6.9672} & 0.8755 & \textcolor{red}{132.0791} & \textcolor{red}{0.8141} & \textcolor{red}{13.9344} \\
\hline
\end{tabular}
\caption{Quantitative comparisons of the 19 pairs of the underexposed and overexposed images by five metrics.}\small
\label{table3}
\end{table}
\begin{figure*}[b]
\centering
\includegraphics[width=0.9\textwidth]{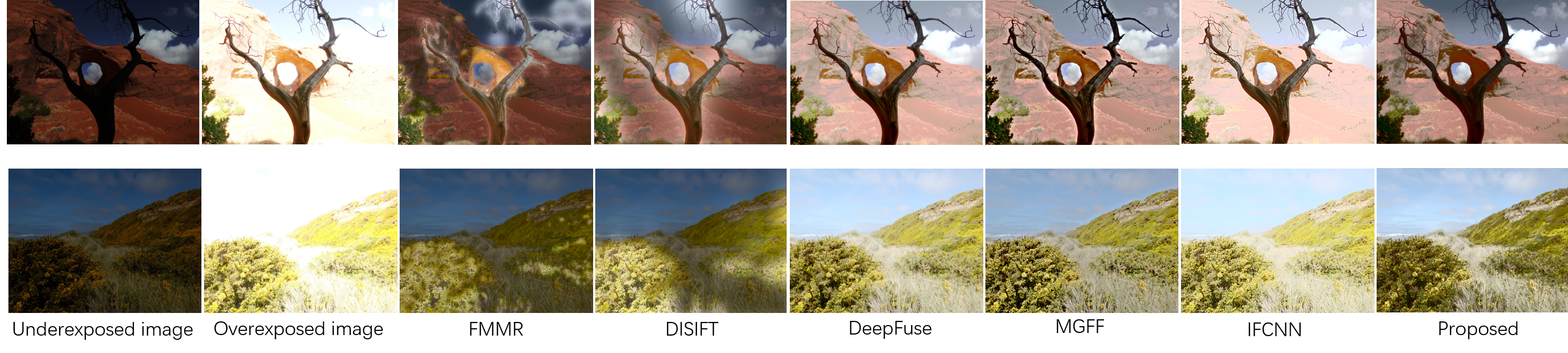} 
\caption{Comparison result of Multi-exposure image fusion. From left to right: underexposed image, overexposed image, the fusion results of FMMR, DISIFT, DeepFuse, MGFF, IFCNN and Proposed.}
\label{fig7}
\end{figure*}

\subsection{Unsupervised MRI and PET image fusion}
In the MRI and PET image fusion experiment, the number of test image pairs is 20. We compare our proposed method with DCHWT\cite{kumar2013multifocus}, Structure Aware\cite{li2018structure}, DDCTPCA\cite{naidu2014hybrid}, PMGI\cite{zhang2020rethinking}, IFCNN\cite{zhang2020ifcnn} by quantitative and qualitative evaluation metrics.
\\ \indent We adopt five related indicators to compare the existing methods, namely entropy (EN), Feature mutual information (FMI), standard deviation (SD),represents the ratio of noise added to the final image(Qabf), and mutual information (MI). As shown in Table ~\ref{table3}, we can observe that the proposed fusion method outperforms in CC, in EN, MI and SD metrics, and the resultant performance is better. We can infer from the results, that the fusion image not only contains the largest amount of information of source multi-modal images but has a strong correlation with each source multi-modal image. Qualitative results can be seen from Figure~\ref{fig7}, the red box highlights the intensity variation of PET colors and local detail in the fused image, it can be seen that the method proposed in this article in the MRI image semantic details of information has very good reservations, at the same time, the larger structure and the functionality of the PET images.
\begin{table}[t]
\renewcommand\tabcolsep{4.5pt}
\centering
\small\begin{tabular}{cccccc}
\hline
Methods & EN & SD & CC & Qabf & MI \\
\hline
DCHWT & \textcolor{red}{5.7501} & 84.9344 & 0.7302 & 0.6849 & \textbf{11.1003} \\
SA & 4.9705 & 84.6888 & 0.7079 & \textcolor{red}{0.7415} & 9.9411 \\
DDCTPCA & 4.8017 & 84.9032 & 0.8030 & 0.4486 & 9.6035 \\
PMGI & 5.3341 & 83.9781 & 0.7646 & \textbf{0.6917} & 10.6684 \\
IFCNN & 4.8187 & \textcolor{red}{98.1931} & \textbf{0.7726} & 0.6662 & 9.6374 \\
Proposed & \textbf{5.3716} & \textbf{94.7850} & \textcolor{red}{0.8998} & 0.2526 & \textbf{10.7433} \\
\hline
\end{tabular}
\caption{Quantitative evaluation for each method via 20 pairs of MRI and PET image fusion.}\small
\label{table3}
\end{table}
\begin{figure*}[t]
\centering
\includegraphics[width=0.9\textwidth]{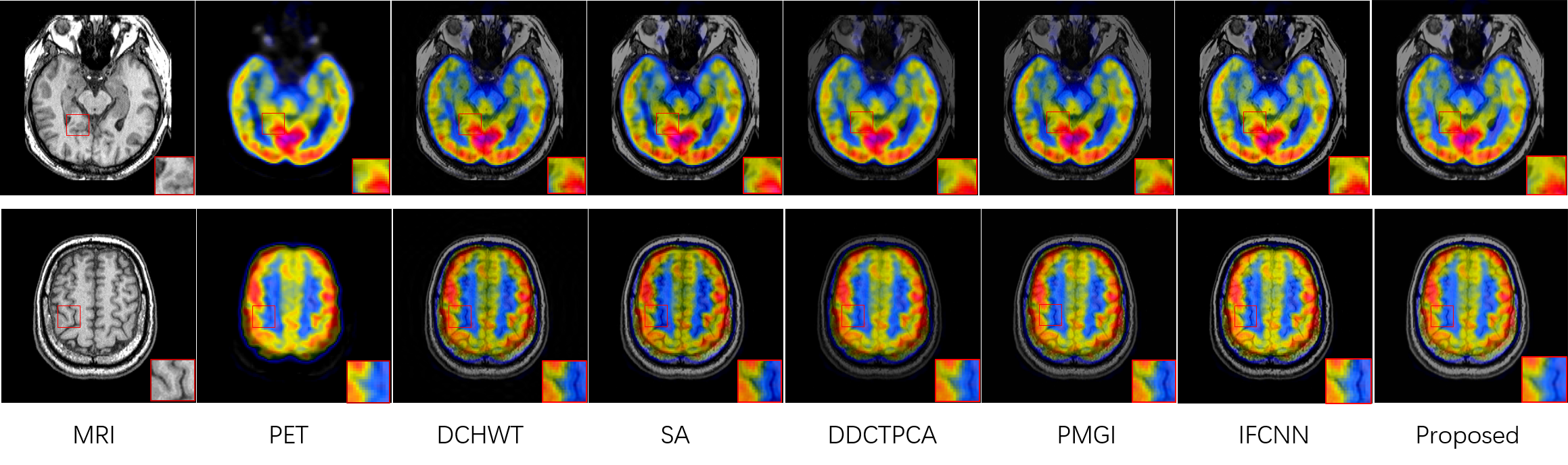} 
\caption{Comparison results of MRI and PET image fusion. From left to right: MRI image, PET image, the results of DCHWT, Structure Aware(SA), DDCTPCA, PMGI, IFCNN and Proposed.}
\label{fig7}
\end{figure*}

\subsection{Unsupervised Multi-focus Image Fusion}
\begin{table}[t]
\renewcommand\tabcolsep{4.5pt}
\centering
\small\begin{tabular}{cccccc}
\hline
Methods & EN & FMI & SD & MI & CC \\
\hline
VSMWLS & 7.4934 & 0.8850 & 106.0445 & 15.0068 & \textbf{0.9564} \\
CBF & 7.4699 & 0.8927 & 106.5535 & 14.9399 & 0.9514 \\
SA & 7.3825 & \textcolor{red}{0.8943} & 107.3123 & 14.7652 & 0.9495 \\
ConvSR & 7.4615 & \textbf{0.8940} & 107.3435 & 14.9230 & 0.9496 \\
MFF & \textbf{7.5666} & 0.8826 & \textbf{111.8743} & \textbf{15.1333} & 0.9515 \\
Proposed & \textcolor{red}{7.6770} & 0.8826 & \textcolor{red}{120.8003} & \textcolor{red}{15.3542} & \textcolor{red}{0.9571} \\
\hline
\end{tabular}
\caption{Quantitative evaluation for each specialized method via 18 pairs of Far-focused and Near-focused image fusion.}\small
\label{table4}
\end{table}
\begin{figure*}[!ht]
\centering
\includegraphics[width=0.9\textwidth]{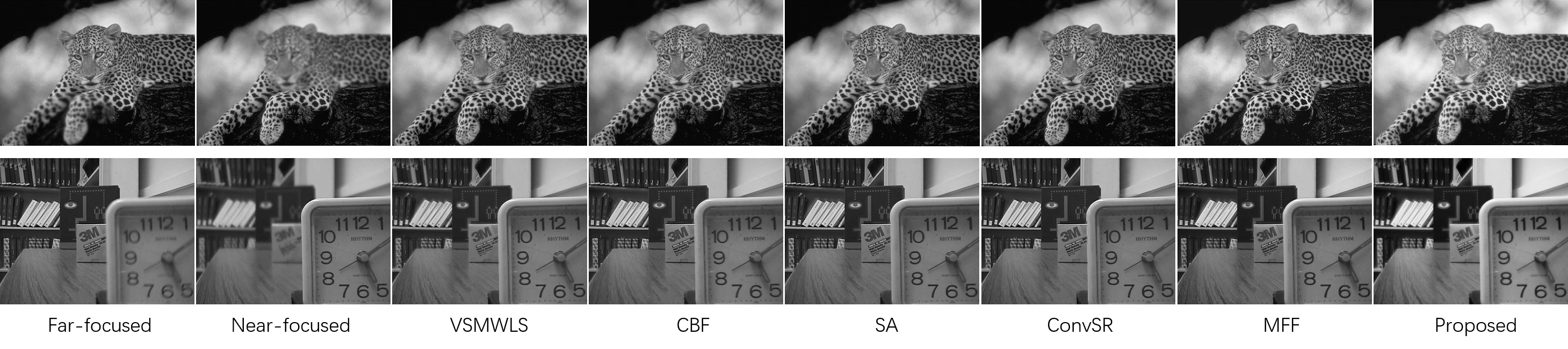} 
\caption{Comparison result of Multi-focus image fusion. From left to right: Far-focused, Near-focused, the results of VSMWLS, CBF, Structure Aware, ConvSR, MFF and Proposed.}
\label{fig8}
\end{figure*}
To verify the proposed method's ability which extracts features, versatility, and generalization from source images, we conducted fusion experiments based on the near-focused and far-focused image datasets. At the same time, our proposed method was compared with other methods, which specialized focused on the far and near-focused images fusion task. We employ five existing methods to compare with our method: VSMWLS\cite{ma2017infrared}, CBF\cite{shreyamsha2015image}, Structure Aware\cite{li2018structure}, ConvSR\cite{liu2016image} and MFF\cite{zhang2021mff}, respectively.
\\ \indent For quantitative analysis, we adopt five related indicators to compare the existing methods, namely entropy (EN), feature mutual information (FMI), standard deviation (SD), mutual information (MI), and correlation coefficient (CC). As shown in Table~\ref{table4}, we can infer that the proposed fusion method outperforms in SD, EN, MI, and CC metric. More intuitively for qualitative analysis, Figure~\ref{fig8}, it demonstrates that our proposed method has a good feature extraction capability of illumination information and spatial details information for multi-focus images compared with the specialized methods.

\subsection{Ablation Study and Visualization}
\begin{table*}[!ht]
\renewcommand\tabcolsep{4.5pt}
\centering
\small\begin{tabular}{cccccc}
\hline
Methods & EN & SCD & SD & Qabf & MI \\
\hline
OD-Net & 6.7359 & 1.4959 & 74.0873 & 0.3704 & 13.4719 \\
CA-Net & 6.9940 & 1.7960 & 80.4371 & 0.3505 & 13.9881 \\
CA-DRDB-Net & 6.7151 & 1.7876 & 72.4742 & 0.3040 & 13.4304 \\
CA-DRDB-AU-Net & 7.2159 & 1.7548 & \textcolor{red}{110.0282} & 0.2747 & 14.4956 \\
CA-DRDB-AU-GRL-Net/Proposed & \textcolor{red}{7.4112} & \textcolor{red}{1.9044} & 103.5565 & \textcolor{red}{0.4070} & \textcolor{red}{14.8226} \\
\hline
\end{tabular}
\caption{Ablation Study on the cross attention-guided network for infrared and visible image fusion.}\small
\label{table4}
\end{table*}
\begin{figure*}[!ht]
\centering
\includegraphics[width=0.9\textwidth]{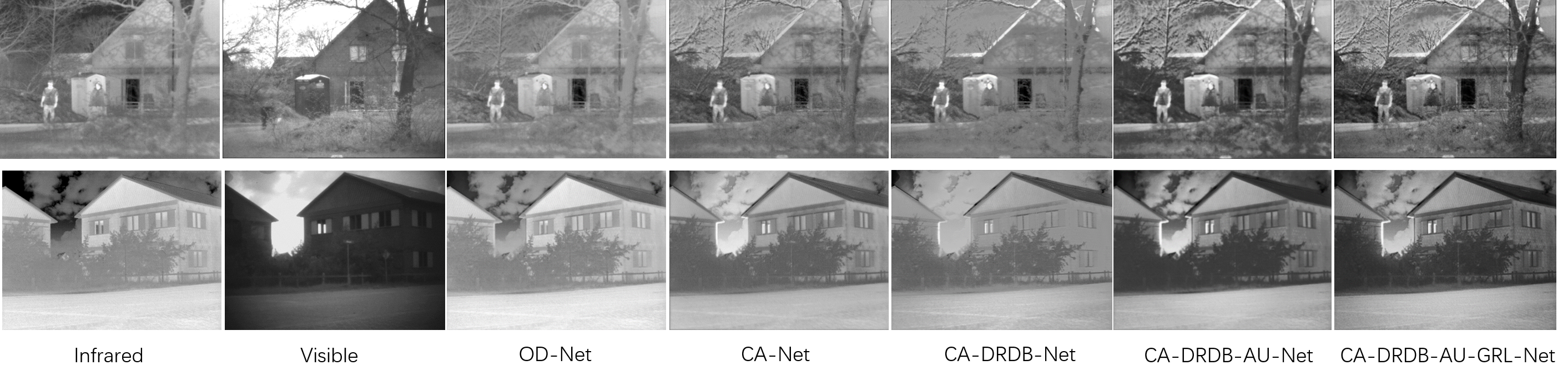} 
\caption{The attention map of cross attention unit from the source images.}
\label{fig8}
\end{figure*}
\begin{figure*}[!ht]
\centering
\includegraphics[width=0.9\textwidth]{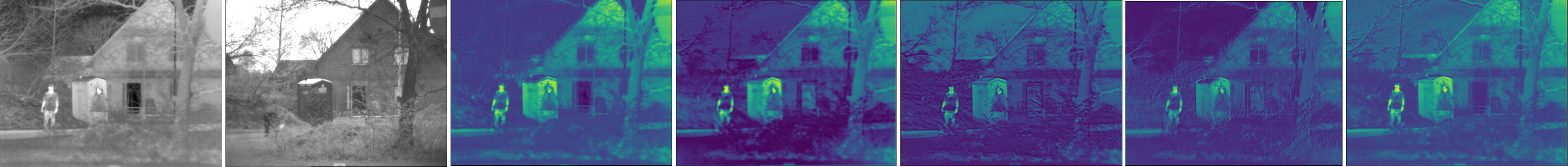} 
\caption{The attention map of cross attention unit from the source images.}
\label{fig9}
\end{figure*}
\subsubsection{Ablation Study}
In this section, we take an ablation study for validating the importance of different components, which is conducted on infrared and visible images fusion experiments as an example. We achieve the ablation study by comparing the following experiments:
\begin{itemize}
\item \textbf{OD-Net}. We take an original densenet(OD-Net) as a baseline model and remove all components, in which the concatenated image from different source images are directly fed to the network. As shown in Figure~\ref{fig8}, the feature information of the fusion image is more likely to come from one source image(Infrared image).
\item \textbf{CA-Net}. Cross-attention(CA-Net) module is effective for modeling the spatial correspondence of source images. As shown in Table~\ref{table4}, compared with OD-Net,  the quantitative indicators are all improved. Through qualitative analysis, details and significant information from the source image are well preserved(See Figure~\ref{fig8}).

\item \textbf{CA-DRDB-Net}. Dilated residual dense block(DRDB) module can obtain sufficient information from the reconstructed image based on a multi-illusion image by use dilated convolution to obtain a larger receptive field to generate illusion and details(See Figure~\ref{fig8}).

\item \textbf{CA-DRDB-AU-Net}. An auxiliary(AU) network for capturing more global context information from the source images. As shown in Table~\ref{table4}, we can observe that the proposed fusion method outperforms in SD metric, which indicates that the quality of the image is more fidelity, and at the same time, other quantitative metrics improved significantly. As shown in Figure~\ref{fig8}, semantic information of the fusion image from one source image is random little more prominent than another source image.
\item \textbf{CA-DRDB-AU-GRL-Net}. In the full model of this paper, we adopt global residual learning to transfer information from the pervious layers. As shown in Table~\ref{table4}, quantitative comparisons of the results outperform, and as displayed in Figure~\ref{fig8}, the result performs better on modeling fusion from source images.
\end{itemize}

\subsubsection{Visualization}
As shown in Figure~\ref{fig9}, we visualize the cross-attention maps obtained by each cross-attention module. Taking the infrared and visible image fusion task as an example, we can infer from the figure that the fusion image background, detail information, and significance features obtained by each attention module are enhanced. It shows that the cross attention-guided is effective for the extraction of spatial correspondence feature information from source images.

\section{Conclusion}
This work introduced a novel unsupervised framework for the challenging multi-image fusion task, and it can serve as a unified framework for four tasks including infrared and visual image fusion, medical image fusion, multi-focus image fusion, and multi-exposure image fusion. To learn semantic features better for image fusion effectively, a cross-attention-guided dense network is proposed to learn the spatial correspondence explicitly, and only the essential details are then learned and aligned attentively. We also proposed an auxiliary branch and a global merging block to better model the long-range relationship and better leverage the global information to reconstruct the fused image. Through quantitative comparison and qualitative analysis, the proposed method achieves better results compared with the state-of-the-art fusion methods.
\section{Acknowledgments}
This work was partially supported by  the Scientific Innovation 2030 Major Project for New Generation of AI under Grant 2020AAA0107300.




\begin{thebibliography}{99}
\footnotesize
\itemsep=-3pt plus.2pt minus.2pt
\baselineskip=13pt plus.2pt minus.2pt
\bibitem{li2018densefuse}
H.~Li and X.-J. Wu, ``Densefuse: A fusion approach to infrared and visible
  images,'' \emph{IEEE Transactions on Image Processing}, 2018, 28(5): 2614--2623. [\textcolor{blue}{DOI: 10.1109/TIP.2020.2977573}]

\bibitem{wang2020deep}
Y.~Wang, W.~Huang, F.~Sun, T.~Xu, Y.~Rong, and J.~Huang, ``Deep multimodal
  fusion by channel exchanging,'' \emph{Advances in Neural Information
  Processing Systems}, 2020, 33.

\bibitem{ma2020ddcgan}
J.~Ma, H.~Xu, J.~Jiang, X.~Mei, and X.-P. Zhang, ``Ddcgan: A dual-discriminator
  conditional generative adversarial network for multi-resolution image
  fusion,'' \emph{IEEE Transactions on Image Processing}, 2020, 29:
  4980--4995. [\textcolor{blue}{DOI: 10.1109/TIP.2020.2977573}]

\bibitem{zhang2021mff}
H.~Zhang, Z.~Le, Z.~Shao, H.~Xu, and J.~Ma, ``Mff-gan: An unsupervised
  generative adversarial network with adaptive and gradient joint constraints
  for multi-focus image fusion,'' \emph{Information Fusion},2021, 66:
  40--53. [\textcolor{blue}{DOI: 10.1016/j.inffus.2020.08.022}]

\bibitem{ram2017deepfuse}
K.~Ram~Prabhakar, V.~Sai~Srikar, and R.~Venkatesh~Babu, ``Deepfuse: A deep
  unsupervised approach for exposure fusion with extreme exposure image
  pairs,'' in \emph{Proceedings of the IEEE international conference on
  computer vision}, 2017, 4714--4722. [\textcolor{blue}{DOI: 10.1109/ICCV.2017.505}]

\bibitem{cao2014multi}
L.~Cao, L.~Jin, H.~Tao, G.~Li, Z.~Zhuang, and Y.~Zhang, ``Multi-focus image
  fusion based on spatial frequency in discrete cosine transform domain,''
  \emph{IEEE signal processing letters}, 2015, 22(2): 220--224. [\textcolor{blue}{DOI: 10.1109/LSP.2014.2354534}]

\bibitem{bin2016efficient}
Y.~Bin, Y.~Chao, and H.~Guoyu, ``Efficient image fusion with approximate sparse
  representation,'' \emph{International Journal of Wavelets, Multiresolution
  and Information Processing}, 2016, 14(04): 1650024. [\textcolor{blue}{DOI: 10.1142/S0219691316500247}]

\bibitem{zhang2018sparse}
Q.~Zhang, Y.~Liu, R.~S. Blum, J.~Han, and D.~Tao, ``Sparse representation based
  multi-sensor image fusion for multi-focus and multi-modality images: A
  review,'' \emph{Information Fusion}, March 2018, 40: 57--75. [\textcolor{blue}{DOI: 10.1016/j.inffus.2017.05.006}]

\bibitem{kuncheva2013pca}
L.~I. Kuncheva and W.~J. Faithfull, ``Pca feature extraction for change
  detection in multidimensional unlabeled data,'' \emph{IEEE transactions on
  neural networks and learning systems}, Jan 2014, 25(1):69--80. [\textcolor{blue}{DOI: 10.1109/TNNLS.2013.2248094}]


\bibitem{ma2019fusiongan}
J.~Ma, W.~Yu, P.~Liang, C.~Li, and J.~Jiang, ``Fusiongan: A generative
  adversarial network for infrared and visible image fusion,''
  \emph{Information Fusion}, August 2019, 48: 11--26. [\textcolor{blue}{DOI: 10.1016/j.inffus.2018.09.004}]

\bibitem{zhang2020rethinking}
H.~Zhang, H.~Xu, Y.~Xiao, X.~Guo, and J.~Ma, ``Rethinking the image fusion: A
  fast unified image fusion network based on proportional maintenance of
  gradient and intensity,'' in \emph{Proceedings of the AAAI Conference on
  Artificial Intelligence}, 2020, 12797--12804. [\textcolor{blue}{DOI: 10.1609/aaai.v34i07.6975}]

\bibitem{xu2020u2fusion}
H.~Xu, J.~Ma, J.~Jiang, X.~Guo, and H.~Ling, ``U2fusion: A unified unsupervised
  image fusion network,'' \emph{IEEE Transactions on Pattern Analysis and
  Machine Intelligence}, 2020. [\textcolor{blue}{DOI: 10.1109/TPAMI.2020.3012548}]


\bibitem{he2016deep}
K.~He, X.~Zhang, S.~Ren, and J.~Sun, ``Deep residual learning for image
  recognition,'' in \emph{Proceedings of the IEEE conference on computer vision
  and pattern recognition}, 2016, pp.770--778. [\textcolor{blue}{DOI: 10.1109/CVPR.2016.90}]

\bibitem{huang2017densely}
G.~Huang, Z.~Liu, L.~Van Der~Maaten, and K.~Q. Weinberger, ``Densely connected
  convolutional networks,'' in \emph{Proceedings of the IEEE conference on
  computer vision and pattern recognition}, 2017, pp.2261--2269. [\textcolor{blue}{DOI: 10.1109/CVPR.2017.243}]

\bibitem{9025565}
A.~A. Bastidas and H.~Tang, ``Channel attention networks,'' in \emph{2019
  IEEE/CVF Conference on Computer Vision and Pattern Recognition Workshops
  (CVPRW)}, 2019, pp.881--888. [\textcolor{blue}{DOI: 10.1109/CVPRW.2019.00117}]

\bibitem{tian2021multi}
M.~Tian, H.~Guo, and C.~Long, ``Multi-level attentive convoluntional neural
  network for crowd counting,'' \emph{arXiv preprint arXiv:2105.11422}, 2021.

\bibitem{song2021cross}
X.~Song, H.~Guo, X.~Xu, H.~Chao, S.~Xu, B.~Turkbey, B.~J. Wood, G.~Wang, and
  P.~Yan, ``Cross-modal attention for mri and ultrasound volume registration,''
  \emph{arXiv preprint arXiv:2107.04548}, 2021.

\bibitem{yan2019attention}
Q.~Yan, D.~Gong, Q.~Shi, A.~v.~d. Hengel, C.~Shen, I.~Reid, and Y.~Zhang,
  ``Attention-guided network for ghost-free high dynamic range imaging,'' in
  \emph{Proceedings of the IEEE/CVF Conference on Computer Vision and Pattern
  Recognition}, 2019, pp.1751--1760. [\textcolor{blue}{DOI: 10.1109/CVPR.2019.00185}]

\bibitem{shreyamsha2015image}
S.K.~BK, ``Image fusion based on pixel significance using cross bilateral filter,'' \emph{Signal, image and video processing}, July 2015, 9(5): 1193--1204. [\textcolor{blue}{DOI: 10.1007/s11760-013-0556-9}]

\bibitem{ma2016infrared}
J.~Ma, C.~Chen, C.~Li and J.~Huang, ``Infrared and visible image fusion via gradient transfer and total variation minimization,'' \emph{Information Fusion}, September 2016, 31: 100--109. [\textcolor{blue}{DOI: 10.1016/j.inffus.2016.02.001}]


\bibitem{li2021rfn}
H.~Li, X.~Wu, J.~Kittler, ``RFN-Nest: An end-to-end residual fusion network for infrared and visible images,'' \emph{Information Fusion}, September 2021, 73: 72--86. [\textcolor{blue}{DOI: 10.1016/j.inffus.2021.02.023}]

\bibitem{zhou2016fusion}
Z.~Zhou, M.~Dong, X.~Xie, and Z.~Gao, ``Fusion of infrared and visible images
  for night-vision context enhancement,'' \emph{Applied optics}, 2016, 55(23): 6480--6490. [\textcolor{blue}{DOI: 10.1364/AO.55.006480}]


\bibitem{naidu2014hybrid}
V.~Naidu, ``Hybrid ddct-pca based multi sensor image fusion,'' \emph{Journal of
  Optics}, March 2014, 43(1): 48--61. [\textcolor{blue}{DOI: 10.1007/s12596-013-0148-7}]

\bibitem{zhang2020ifcnn}
Y.~Zhang, Y.~Liu, P.~Sun, H.~Yan, X.~Zhao, and L.~Zhang, ``Ifcnn: A general
  image fusion framework based on convolutional neural network,''
  \emph{Information Fusion}, February 2020, 54: 99--118. [\textcolor{blue}{DOI: 10.1016/j.inffus.2019.07.011}]


\bibitem{li2020mdlatlrr}
H.~Li, X.-J. Wu, and J.~Kittler, ``Mdlatlrr: A novel decomposition method for
  infrared and visible image fusion,'' \emph{IEEE Transactions on Image
  Processing}, 2020, 29: 4733-4746. [\textcolor{blue}{DOI: 10.1109/TIP.2020.2975984}]



\bibitem{aslantas2015new}
V.~Aslantas and E.~Bendes, ``A new image quality metric for image fusion: the
  sum of the correlations of differences,'' \emph{Aeu-international Journal of
  electronics and communications}, 2015, 69(12): 1890--1896. [\textcolor{blue}{DOI: 10.1016/j.aeue.2015.09.004}]

\bibitem{li2018structure}
W.~Li, Y.~Xie, H.~Zhou, Y.~Han, and K.~Zhan, ``Structure-aware image fusion,''
  \emph{Optik}, November 2018, 172: 1--11. [\textcolor{blue}{DOI: 10.1016/j.ijleo.2018.06.123}]

\bibitem{li2012fast}
S.~Li and X.~Kang, ``Fast multi-exposure image fusion with median filter and
  recursive filter,'' \emph{IEEE Transactions on Consumer Electronics},
  May 2012, 58(2): 626--632. [\textcolor{blue}{DOI: 10.1109/TCE.2012.6227469}]


\bibitem{hayat2019ghost}
N.~Hayat and M.~Imran, ``Ghost-free multi exposure image fusion technique using
  dense sift descriptor and guided filter,'' \emph{Journal of Visual
  Communication and Image Representation},  July 2019, 62: 295--308. [\textcolor{blue}{DOI: 10.1016/j.jvcir.2019.06.002}]


\bibitem{bavirisetti2019multi}
D.~P. Bavirisetti, G.~Xiao, J.~Zhao, R.~Dhuli, and G.~Liu, ``Multi-scale guided
  image and video fusion: A fast and efficient approach,'' \emph{Circuits,
  Systems, and Signal Processing}, December 2019, 38(12): 5576--5605. [\textcolor{blue}{DOI: 10.1007/s00034-019-01131-z}]


\bibitem{kumar2013multifocus}
B.~S. Kumar, ``Multifocus and multispectral image fusion based on pixel
  significance using discrete cosine harmonic wavelet transform,''
  \emph{Signal, Image and Video Processing}, November 2013, 7(6): 1125--1143. [\textcolor{blue}{DOI: 10.1007/s11760-012-0361-x}]


\bibitem{ma2017infrared}
J.~Ma, Z.~Zhou, B.~Wang, and H.~Zong, ``Infrared and visible image fusion based
  on visual saliency map and weighted least square optimization,''
  \emph{Infrared Physics \& Technology}, May 2017, 82: 8--17. [\textcolor{blue}{DOI: 10.1016/j.infrared.2017.02.005}]


\bibitem{liu2016image}
Y.~Liu, X.~Chen, R.~K. Ward, and Z.~J. Wang, ``Image fusion with convolutional
  sparse representation,'' \emph{IEEE signal processing letters}, Dec. 2016, 23(12): 1882--1886. [\textcolor{blue}{DOI: 10.1109/LSP.2016.2618776}]




\end{thebibliography}
\end{document}